\documentclass[runningheads]{llncs}

\usepackage{graphicx}
\usepackage{amsmath}
\usepackage{float}
\usepackage{a4wide}
\usepackage{algorithm}
\usepackage{algorithmicx}
\usepackage{algpseudocode}
\usepackage{epstopdf}
\usepackage{bm}
\usepackage{tikz}
\usepackage{multirow}
\usepackage{amsmath}
\usetikzlibrary{bayesnet}
\usetikzlibrary{shapes.gates.logic.US,trees,positioning,arrows}
\usepackage{caption}
\usepackage{subcaption}
\usetikzlibrary{trees}
\usepackage{mathtools}
\usepackage{amsmath}
\usepackage{booktabs}
\usepackage{rotating}

\usepackage{epstopdf}
\usepackage{amssymb}
\usepackage{microtype}
\usepackage{url}
\usepackage{caption}
\usepackage{subcaption}
\newcommand{\mRVMa}{{mRVM$_1$}}
\newcommand{\mRVMb}{{mRVM$_2$}}

\usepackage[T1]{fontenc}
\usepackage{lmodern}

\begin{document}

	\title{Improving decision-making via risk-based active learning: Probabilistic discriminative classifiers}
	
	%
	%
	\author{A.J.\ Hughes\inst{1} \and
		P.\ Gardner\inst{1} \and L.A.\ Bull\inst{2} \and N.\ Dervilis \inst{1} \and K.\ Worden\inst{1}}
	\authorrunning{A.J.\ Hughes et al.}
	%
	\institute{Dynamics Research Group, Department of Mechanical Engineering, University of Sheffield, Sheffield, \\ S1 3JD, UK
		 \and
		The Alan Turing Institute, The British Library, 96 Euston Road, London, NW1 2DB, UK\\
	\email{aidan.j.hughes@sheffield.ac.uk}}
	\maketitle              
	
	\begin{abstract} 
	
	Gaining the ability to make informed decisions on operation and maintenance of structures provides motivation for the implementation of structural health monitoring (SHM) systems. However, descriptive labels for measured data corresponding to health-states of the monitored system are often unavailable. This issue limits the applicability of fully-supervised machine learning paradigms for the development of statistical classifiers to be used in decision-support in SHM systems. One approach to dealing with this problem is risk-based active learning. In such an approach, data-label querying is guided according to the expected value of perfect information for incipient data points. For risk-based active learning in SHM, the value of information is evaluated with respect to a maintenance decision process, and the data-label querying corresponds to the inspection of a structure to determine its health state.

	In the context of SHM, risk-based active learning has only been considered for generative classifiers. The current paper demonstrates several advantages of using an alternative type of classifier -- discriminative models. Using the Z24 Bridge dataset as a case study, it is shown that discriminative classifiers have benefits, in the context of SHM decision-support, including improved robustness to sampling bias, and reduced expenditure on structural inspections.

	\end{abstract}

	\section{Introduction}  

	Statistical classifiers are a fundamental technology for structural health monitoring (SHM). In general, classification models provide a mapping $f$ from an input space $X \in \mathbb{R}^D$ to a discrete output space $Y \in \{1,\ldots,K\}$, i.e.\ $f: X \rightarrow Y$. In the context of SHM, the input space $X$ represents some discriminative features extracted from data acquired from the monitoring system, and the output space corresponds to labels providing information regarding structural health states. These classification models play a pivotal role in SHM as the predictions that they provide, $p(y=k|\mathbf{x})$, allow for data-informed decision-making regarding the operation and maintenance of high-value or safety-critical structures \cite{Hughes2021,Kamariotis2022,Vega2020a}.

	Unfortunately, for SHM applications, the traditional \textit{supervised} approaches to machine learning are unsuitable for developing the necessary classification models, since comprehensive labelled datasets are seldom available prior to the implementation of a monitoring system. Supervised learning is therefore rendered inapplicable as it requires completely labelled data. Unsupervised learning, can be applied; however, the learned model will be of limited use for decision-making without the contextual information provided by labels.

	In previous works, online algorithms that circumvent the drawbacks of supervised and unsupervised learning were proposed \cite{Bull2019,Hughes2022}. These approaches adopt a form of partially-supervised learning known as \textit{active learning}. Active-learning algorithms build labelled datasets on which to train classifiers in a supervised way by querying label information for otherwise unlabelled incipient data points. In \cite{Hughes2022}, a decision-theoretic measure was used such that the classification models were learned specifically for use in decision-support systems. This decision-centric approach to learning classifiers is termed risk-based active learning.
	
	Previous works presented in \cite{Bull2019,Hughes2022} focussed on learning generative classification models, in particular, Gaussian mixture models. A drawback of generative models, however, is that a distribution must be selected and assumed to be representative of the data. Furthermore, classification models, in general, rely on the assumption that training data are independent and identically distributed (iid). In active learning, both of these assumptions is often violated because of the preferential nature of the label-querying process and results in degradation of performance. This phenomenon is known as \textit{sampling bias}.

	The current paper aims to improve SHM decision-making performance by leveraging characteristics of an alternative form of classifier within the risk-based active learning framework; specifically, the \textit{multiclass relevance vector machine} (mRVM). Importantly, the \textit{discriminative} nature of the mRVM means that the model does not rely on assumptions regarding the underlying distribution of data. Additionally, the \text{sparse} nature of the mRVM means that an effective classifier can be learned using very little data. These characteristics of the mRVM result in improved decision-making performance by reducing the number of inspections made throughout a monitoring campaign and by reducing the number of erroneous decisions made. This is demonstrated in the current paper with use of the Z24 Bridge dataset, following sections that provide background theory on risk-based active learning and mRVMs. Finally, concluding statements are provided.

 \section{Risk-based Active Learning}
	
	Active learning is a form of partially-supervised learning. Partially supervised learning is characterised by the utilisation of both labelled and unlabelled data.

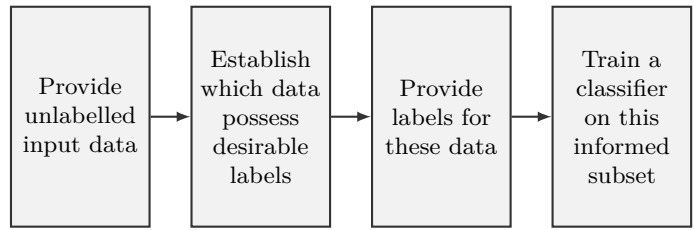
\begin{figure}[ht!]
	\centering
	\begin{tikzpicture}[auto]
		\begin{footnotesize}
			\tikzstyle{block} = [rectangle, thick, draw=black!80, text width=5em, text centered, minimum height=9em, fill=black!5]
			\tikzstyle{line} = [draw, -latex, thick]
			\node [block, node distance=24mm] (A) {Provide\\ unlabelled input data};
			\node [block, right of=A, node distance=24mm] (B) {Establish which data possess desirable labels};
			\node [block, right of=B, node distance=24mm] (C) {Provide labels for these data};
			\node [block, right of=C, node distance=24mm] (D) {Train a classifier on this informed subset};
			\path [line, draw=black!80] (A) -- (B);
			\path [line, draw=black!80] (B) -- (C);
			\path [line, draw=black!80] (C) -- (D);
		\end{footnotesize}
	\end{tikzpicture}
	\caption{A general active learning heuristic.}
	\label{fig:AL1}
\end{figure}

Active learning algorithms construct a labelled data set $\mathcal{D}_l$ by querying labels for otherwise unlabelled data $D_u$. Labels for incipient data are queried preferentially according to some measure of their desirability. Models are then trained in a supervised manner using $\mathcal{D}_l$. A general heuristic for active learning is presented in Figure \ref{fig:AL1}.

In \cite{Bull2019}, an online active learning approach is shown to overcome the primary challenge for classifier development in the context of SHM - initial scarcity of comprehensive labelled data. This result is achieved through the construction of a labelled dataset via a querying process that corresponds to the inspection of a structure by an engineer to determine its health state. In \cite{Hughes2022}, an alternative risk-based approach to active learning is proposed in which \textit{expected value of perfect information} (EVPI) is used as a measure to guide querying.

The expected value of perfect information (EVPI) can be interpreted as the price that a decision-making should be willing to pay in order to gain access to perfect information of an otherwise unknown or uncertain state. Here, it should be clarified that the terminology `perfect information' refers to ground-truth states observed without uncertainty. More formally, the expected value of observing $y$ with perfect information prior to making decision $d$ is given by,

\begin{equation}\label{eq:EVPI}
	\text{EVPI}(d | y) := \text{MEU}(\mathcal{I}_{y \rightarrow d}) - \text{MEU}(\mathcal{I})
\end{equation}

\noindent
where $\mathcal{I}$ is an influence diagram representing an SHM decision process with decision $d$ influenced by random variable $y$, and $\mathcal{I}_{y \rightarrow d}$ is a modified influence diagram incorporating an additional informational link from $y$ to $d$ indicating that $y$ is observed prior to $d$. An example calculation of EVPI is presented in \cite{Hughes2022}. Details of the procedure for forming SHM decision processes as influence diagrams can be found in \cite{Hughes2021}.

A convenient criterion for mandating structural inspections can be derived from the EVPI. Simply, if the EVPI of a label $y_t$ for a data point $\tilde{\mathbf{x}}_t$ exceeds the cost of making an inspection $C_{\text{ins}}$ then the ground-truth for $y_t$ should be obtained prior to $d_t$. Subsequently, the labelled dataset $\mathcal{D}_l$ can be extended to include the newly acquired data-label pair $(\mathbf{x}_t,y_t)$, and the classifier retrained. While the assumption of perfect information may not hold in all cases, the principles and methodologies discussed in the current paper hold in general for value of information. Furthermore, the perfect information assumption may be relaxed via the introduction of an additional probabilistic model that quantifies uncertainty in inspections. Full details of the risk-based active learning algorithm can be found in \cite{Hughes2022}.

Adopting a risk-based approach to active learning allows one to learn statistical classifiers with consideration for the decision support contexts in which they may be employed. In \cite{Hughes2022}, it is demonstrated that this approach provides a cost-efficient manner for classifier development.

	\section{Probabilistic Discriminative Classifiers}
	
	Probabilistic discriminative models (sometimes referred to as conditional models), provide an alternative approach to generative models for developing statistical classifiers. Whereas generative models seek to first represent the underlying joint probability distribution $p(\mathbf{x},y)$, discriminative models seek to learn the predictive conditional distribution $p(y | \mathbf{x})$ directly. For discriminative classifiers, the mapping $f: X \rightarrow Y$ is typically specified via boundaries that partition the feature space $X$ according to $Y$. In probabilistic discriminative models, these classification boundaries are `soft', allowing for nondeterministic classification. Discriminative models do not necessarily require training datasets to be representative of the underlying distribution. Because of this property, discriminative models are more robust to sampling bias, thus are considered as a candidate approach for improving risk-based active learning.

	For its sparsity and robust uncertainty quantification, a Bayesian model known as the \textit{relevance vector machine} (RVM) is selected as the probabilistic discriminative classifier for the current paper.

	\subsection{Multiclass Relevance Vector Machines} 
	 
	Originally introduced by Tipping in \cite{Tipping2001}, the RVM is a computationally-efficient Bayesian model capable of achieving high accuracies for both regression and classification tasks via the use of a sparse subset of training data. For the multiclass classification of an unlabelled data point $\tilde{\mathbf{x}}_t$, mRVMs employ a fundamentally linear model-form,
	
	\begin{equation}
		\mathbf{f}_t = \mathbf{W}^{\top}\mathbf{k}(\mathbf{X}_l,\tilde{\mathbf{x}}_t)
	\end{equation}
	
	\noindent where $\mathbf{f}_t = \{f_1,\ldots,f_K\}^{\top}$ is a vector of $K$ auxiliary variables that provide a ranking system by which the class membership of an unlabelled data point may be assessed. $\mathbf{k}(\mathbf{X}_l,\tilde{\mathbf{x}}_t)$ is an $n \times 1$ vector for which the $i^{\text{th}}$ element is specified by the \textit{kernel function} $k(\mathbf{x}_i,\tilde{\mathbf{x}}_t)$. The kernel function specifies a set of basis functions that reflect the similarity between $\tilde{\mathbf{x}}_t$ and training inputs $\mathbf{X}_l = \{\mathbf{x}_i|(\mathbf{x}_i,y_i) \in \mathcal{D}_l \}^n_{i=1}$; nonlinearity can be introduced into the RVM by selecting a nonlinear kernel function. $\mathbf{W} = \{ \mathbf{w}_1, \ldots,\mathbf{w}_K\}$ is an $n \times K$ matrix of tunable parameters referred to as \textit{weights}, and where $\mathbf{w}_k = \{w_{1,k},\ldots,w_{n,k}\}^{\top}$. These weights act as a voting system that indicate which data in $\mathcal{D}_l$ are important, or `relevant', for discriminating between classes.

	 Predicted class labels $\hat{y}_t$ are related to otherwise unlabelled data via the auxiliary variables $\mathbf{f}_t$ by according to the multinomial probit link and the corresponding multinomial probit likelihood \cite{Damoulas2008}, given by,
	
	\begin{equation}\label{eq:RVMLabel1}
		\hat{y}_t = k \iff f_k > f_j \text{ } \forall j \neq k
	\end{equation}
	
	\noindent and,
	
	\begin{equation}\label{eq:multi_probit_lik}
		p(\tilde{y}_t = k | \mathbf{W},\mathbf{k}) = \mathbb{E}_{p(u)} \Biggl[ \prod_{j \neq k} \Phi (u + (\mathbf{w}_k - \mathbf{w}_{j})^{\top}\mathbf{k}) \Biggr]
	\end{equation}
	
	\noindent respectively. Further details relating to the likelihood, priors, and inference algorithms used for mRVMs can be found in \cite{Psorakis2010}.
	
	RVMs are able to achieve high computational efficiency by using a subset of $n^{\ast}$ `relevant' samples from $\mathcal{D}_l$ such that $n^{\ast} << n$. This reduced subset is denoted $\mathcal{A}$. Two approaches for achieving this sparsity are presented in \cite{Psorakis2010} and termed \mRVMa{} and \mRVMb{}.
	
	\mRVMa{} utilises a constructive approach to build $\mathcal{A}$. $\mathcal{A}$ is initialised as an empty set with samples subsequently added or removed from $\mathcal{A}$ based upon their contribution to the objective function given by the decomposed form of the marginal log-likelihood detailed in \cite{Psorakis2010}. Whereas \mRVMa{} provides a constructive approach to the formation of $\mathcal{A}$ of relevance vectors, the approach termed \mRVMb{} sculpts $\mathcal{A}$ from $\mathcal{D}_l$ by iteratively discarding samples with scales $\alpha_{i,k}$ sufficiently large such that $w_{i,k}$ is negligible. Further details and characteristics of the \mRVMa{} and \mRVMb{} algorithms are provided in \cite{Psorakis2010}.
	
	By utilising the sparse subset $\mathcal{A}$ to form the basis functions of the mRVM classification model, improvements in computational efficiency are achieved. Furthermore, in the context of risk-based active learning, it is hypothesised that probabilistic discriminative classifiers will show robustness to sampling bias over the generative classifiers -- in part because detrimental or superfluous data are excluded from the model by virtue of its sparsity, and because discriminative classifiers, in general, do not rely upon assumptions regarding the underlying distribution of the data.

	\section{Case Study: Z24 Bridge}
	
	In the current section, the benefits of utilising probabilistic discriminative classifiers within a risk-based active learning framework are assessed using an experimental dataset obtained from the Z24 Bridge \cite{Maeck2003}.
	
	The Z24 bridge was a concrete highway bridge in Switzerland, situated between the municipalities of Koppigen and Utzenstorf and near the town of Solothurn. The bridge was the subject of a cross-institutional research project (SIMCES). The purpose of this project was to generate a benchmark dataset and prove the feasibility of vibration-based SHM \cite{Maeck2001,DeRoeck2003}. Because of the presence of varying structural and environmental conditions, the benchmark dataset has seen wide use in research focussed on SHM and modal analysis. Prior to its demolition, the Z24 bridge was instrumented with sensors, and for a period of 12 months, the dynamic response of the structure and the environmental conditions were monitored. The quantities measured included the accelerations at multiple locations, air temperature, deck temperature, and wind speed \cite{Peeters2001}.
	
	Using the dynamic response data obtained during the monitoring campaign, the first four natural frequencies of the structure were obtained. A visualisation of these natural frequencies can be seen in Figure \ref{fig:all_queries_rvm_z24}.
	
	In total, the dataset consists of 3932 observations. Towards the end of the monitoring campaign -- from observation 3476 onwards -- incremental damage was introduced to the structure. Throughout the campaign, the bridge experienced low temperatures, and as such, the structure exhibited cold temperature effects, particularly noticeable between observations 1200 and 1500. It is believed that the increased natural frequencies observed during these periods are a result of stiffening in the bridge deck induced by very low ambient temperatures. From these data, a four-class classification problem can be defined such that $y_t \in \{ 1,2,3,4 \}$:
	
	\begin{itemize}
		\item Class 1: normal undamaged condition (green)
		\item Class 2: cold temperature undamaged condition (blue)
		\item Class 3: incipient damage condition (orange)
		\item Class 4: advanced damage condition (pink).
	\end{itemize}
	
	In accordance with \cite{Hughes2022}, here, it is assumed that the data obtained after the introduction of damage can be separated into two halves; the earlier half corresponding to incipient damage and the later half corresponding to advanced damage. This assumption is deemed to be reasonable because of the incremental nature of the damage progression \cite{Maeck2003}.
	
	\subsection{Decision Process}
	
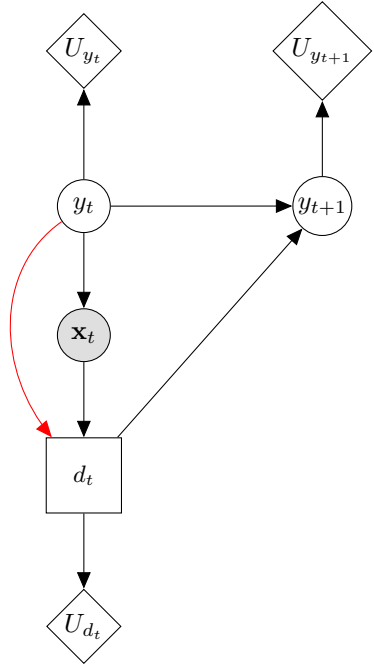
\begin{figure}[ht!]
		\centering
		\begin{tikzpicture}[x=1.7cm,y=1.8cm]
	
			\node[det] (uf1) {$U_{y_{t}}$} ;
			\node[det, right=2cm of uf1] (uf2) {$U_{y_{t+1}}$} ;
			\node[latent, below=1.2cm of uf1] (x1) {$y_{t}$} ;
			\node[latent, below=1cm of uf2] (x2) {$y_{t+1}$} ;
			\node[obs, below=1cm of x1] (y1) {$\mathbf{x}_{t}$} ;
			\node[rectangle,draw=black,minimum width=1cm,minimum height=1cm,below=1cm of y1] (d1) {$d_{t}$} ;
			\node[det, below=1cm of d1] (u1) {$U_{d_{t}}$} ;
	
			\edge {x1} {uf1} ; %
			\edge {x2} {uf2} ; %
			\edge {x1} {x2} ; %
			\edge {x1} {y1} ; %
			\edge {d1} {x2} ; %
			\edge {d1} {u1} ; %
			\edge {y1} {d1} ; %
	
			\draw [red, ->] (x1) to [out=-145,in=130] (d1);

		\end{tikzpicture}
		\caption{An influence diagram of a simplified SHM maintenance decision. The red edge denoting inspection prior to maintenance (shown in red) is present in $\mathcal{I}_{y_t \rightarrow d_t}$ and absent in $\mathcal{I}$.}
		\label{fig:OverallPGM2}
	\end{figure}

	An operation and maintenance (O\&M) decision process must be specified for the Z24 Bridge dataset in order to apply a risk-based active learning approach to classifier development. As the bridge was demolished over two decades ago at the time of writing, here, a simple binary decision process (`do nothing' or `perform maintenance') is considered. The graphical representation of the decision process is provided by the influence diagram in Figure \ref{fig:OverallPGM2}. The parameters defining the conditional probability distributions and utility functions given by the influence diagram are specified as follows with a discussion of the assumptions used to specify these parameters provided in \cite{Hughes2022}.

	Given the decided action $d_t = 0$, the health-state transitions are specified such that the structure monotonically degrades, with a propensity to remain in the current state. States 1 and 2 both correspond to an undamaged condition of the bridge and reflect differing environmental conditions. As such, transitions both to and from these states must be possible according to the transition model. Therefore, it is simply asserted that if there was a cold temperature for the most recent measurement, it is probable that the next measurement will also be made at a cold temperature. Likewise, the same reasoning is applied to the normal condition data. These assumptions are represented in the conditional probability distribution shown in Table \ref{tab:P_y1_y0_d0_z24}.

	The portion of the transition model for $d_t = 1$ is presented in Table \ref{tab:P_y1_y0_d1_z24}. This distribution is specified via the assumption that the act of performing maintenance returns the bridge to an undamaged condition with high probability. Since there are two states corresponding to the undamaged condition of the bridge, the transition model must be specified such that the probability of returning to each of the undamaged condition states is independent of the action taken, i.e.\ it must be asserted that the weather condition is not influenced by the action taken. The procedure by which this constraint is imposed is outlined in \cite{Hughes2022}.

	\begin{table}
		\begin{minipage}{.5\linewidth}
		  \centering
		  \caption{The conditional probability \\ table $P(y_{t+1}|y_t, d_t)$ for $d_t = 0$.}
		\label{tab:P_y1_y0_d0_z24}       
		\begin{tabular}{c c c c c c}
			\toprule
			\midrule
			& & \multicolumn{4}{c}{$y_{t+1}$}\\
			&& 1  & 2 & 3 & 4  \\ \cmidrule{3-6}
			\multicolumn{1}{c}{\multirow{4}{*}{\begin{sideways}\parbox{1.5cm}{\centering $y_t$}\end{sideways}}}   &
			\multicolumn{1}{l}{1}& 0.7 & 0.28 & 0.015 & 0.005 \\
			\multicolumn{1}{c}{}    &
			\multicolumn{1}{l}{2}& 0.43 & 0.55 & 0.15 & 0.05  \\
			\multicolumn{1}{c}{}    &
			\multicolumn{1}{l}{3} & 0 & 0 & 0.8 & 0.2  \\
			\multicolumn{1}{c}{}    &   
			\multicolumn{1}{l}{4} & 0 & 0 & 0 & 1  \\
			\midrule
			\bottomrule
		\end{tabular}
		\end{minipage}%
		\begin{minipage}{.5\linewidth}
		  \centering
		  \caption{The conditional probability \\ table $P(y_{t+1}|y_t, d_t)$ for $d_t = 1$.}
		\label{tab:P_y1_y0_d1_z24}       
		\begin{tabular}{c c c c c c}
			\toprule
			\midrule
			& & \multicolumn{4}{c}{$y_{t+1}$}\\
			&& 1  & 2 & 3 & 4  \\ \cmidrule{3-6}
			\multicolumn{1}{c}{\multirow{4}{*}{\begin{sideways}\parbox{1.5cm}{\centering $y_t$}\end{sideways}}}   &
			\multicolumn{1}{l}{1}& 0.7173 & 0.2857 & 0 & 0 \\
			\multicolumn{1}{c}{}    &
			\multicolumn{1}{l}{2}& 0.4388 & 0.5612 & 0 & 0  \\
			\multicolumn{1}{c}{}    &
			\multicolumn{1}{l}{3} & 0.5996 & 0.3904 & 0.01 & 0  \\
			\multicolumn{1}{c}{}    &   
			\multicolumn{1}{l}{4} & 0.5996 & 0.3904 & 0 & 0.01  \\
			\midrule
			\bottomrule
		\end{tabular}
		\end{minipage}
	  \end{table}

	For brevity, it is assumed that utilities can be attributed directly to the four health states comprising the Z24 dataset, without the need for a fault tree. Here States 1 and 2 are assigned some small positive utility as reward for the bridge being functional, with minimal risk of failure. A methodology for relating costs to health states via fault tree representations of failure modes is provided in \cite{Hughes2021}. The incipient damage state is assigned a negative utility of moderate magnitude. This value is to reflect a possible reduction in operating capacity (e.g.\ bridge closure), and low-to-moderate risk of failure. Class 4, which corresponds to the advanced damage condition, is assigned a very large negative utility. This value is to reflect the severe consequences associates with the collapse of a bridge. The utility function $U(y_t)$ is provided in Table \ref{tab:Uy_z24}.

	The utilities associated with the candidate actions in the domain of $d_t$ are given by the utility function $U(d_t)$ and are presented in Table \ref{tab:Ud_z24}. It is assumed that the action $d_t=0$ (`do nothing') has zero utility and that the action $d_t=1$ (`perform maintenance') has negative utility to reflect the resource expenditure on repairing a bridge.

	\begin{table}
		\begin{minipage}{.5\linewidth}
		  \centering
		  \caption{The utility function $U(y_{t+1})$.}
		  \label{tab:Uy_z24}   
		  \begin{tabular}{cc}
			  \toprule
			  \midrule
			  $y_{t+1}$ & $U(y_{t+1})$\\
			  \midrule
			  $1$ & $10$\\
			  $2$ & $10$\\
			  $3$ & $-50$\\
			  $4$ & $-1000$\\
			  \midrule
			  \bottomrule
		  \end{tabular}
		\end{minipage}%
		\begin{minipage}{.5\linewidth}
		  \centering
		  \caption{The utility function $U(d_t)$ where $d_t=0$ and $d_t=1$ denote the `do nothing' and `repair' actions, respectively.}
		\label{tab:Ud_z24}   
		\begin{tabular}{cc}
			\toprule
			\midrule
			$d_t$ & $U(d_t)$\\
			\midrule
			$0$ & $0$\\
			$1$ & $-100$\\
			\midrule
			\bottomrule
		\end{tabular}
		\end{minipage}
	  \end{table}
	
	Finally, the cost of inspection is specified to be $C_{\text{ins}} = 30$. The moderate cost is intended to reflect resources required to inspect a large-scale structure such as the Z24 Bridge.

\subsection{Results}

Discriminative classifiers, in the form of \mRVMa{} and \mRVMb{}, were used to provide the probability distributions $p(y_t|\mathbf{x}_t)$ for decision processes as specified in the previous section. Risk-based active learning was employed to develop these classification models for the Z24 Bridge dataset. For comparison with the probabilistic discriminate classifiers, risk-based active learning of a Gaussian mixture model was also conducted; details of which can be found in \cite{Hughes2022}. 

To assess the decision-making performance achieved via each classifier, a quantity termed `decision accuracy' is used. The decision accuracy is a comparison between the actions selected by an agent using the classifier being evaluated, and the optimal (correct) actions selected by an agent in possession of perfect information \cite{Hughes2021}.

\begin{figure}[ht!]
	\begin{subfigure}{.5\textwidth}
		\centering
		\scalebox{0.4}{
			\includegraphics{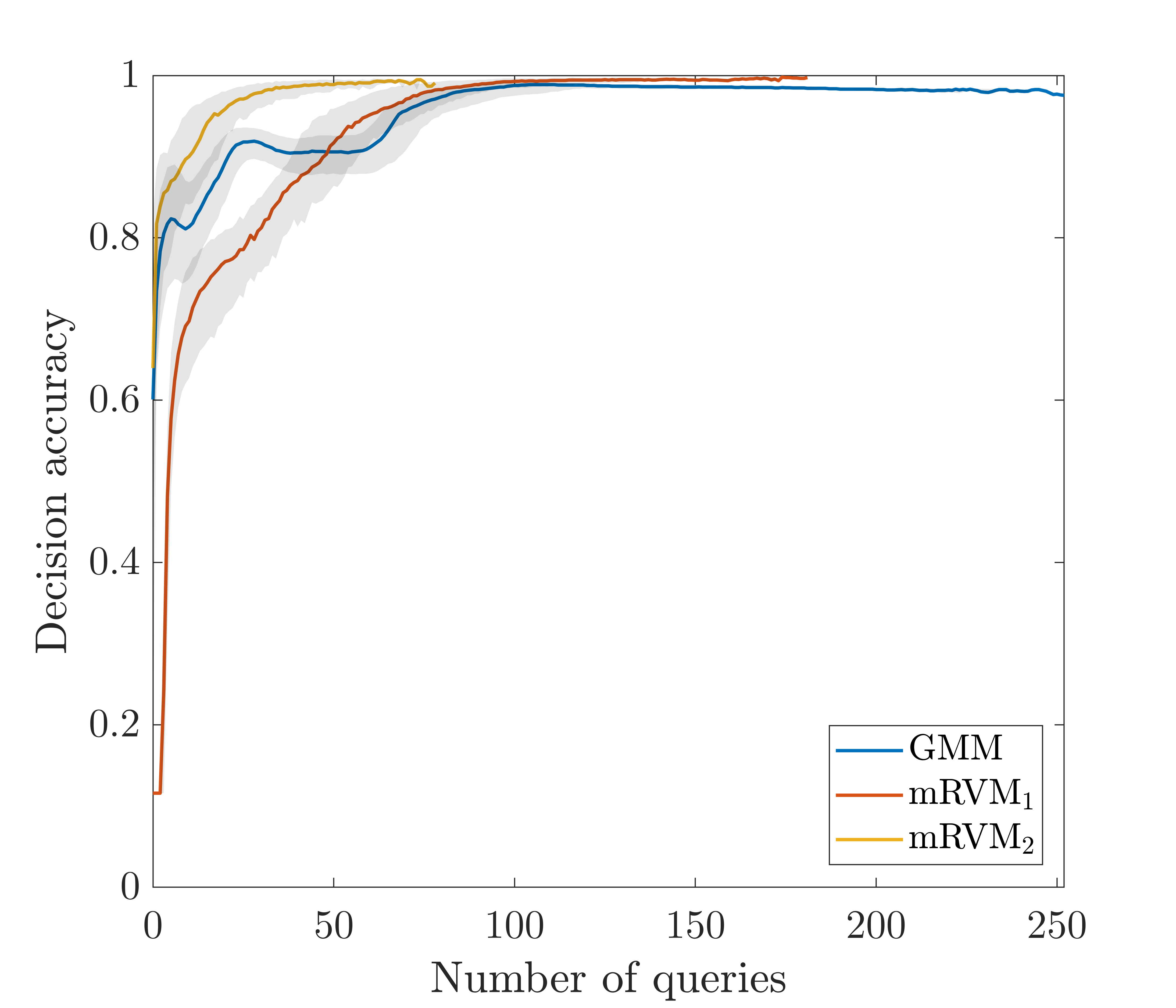}
		}
		\caption{ }
		\label{fig:dacc_rvm_z24} 
	\end{subfigure}
	\begin{subfigure}{.5\textwidth}
		\centering
		\scalebox{0.4}{
			\includegraphics{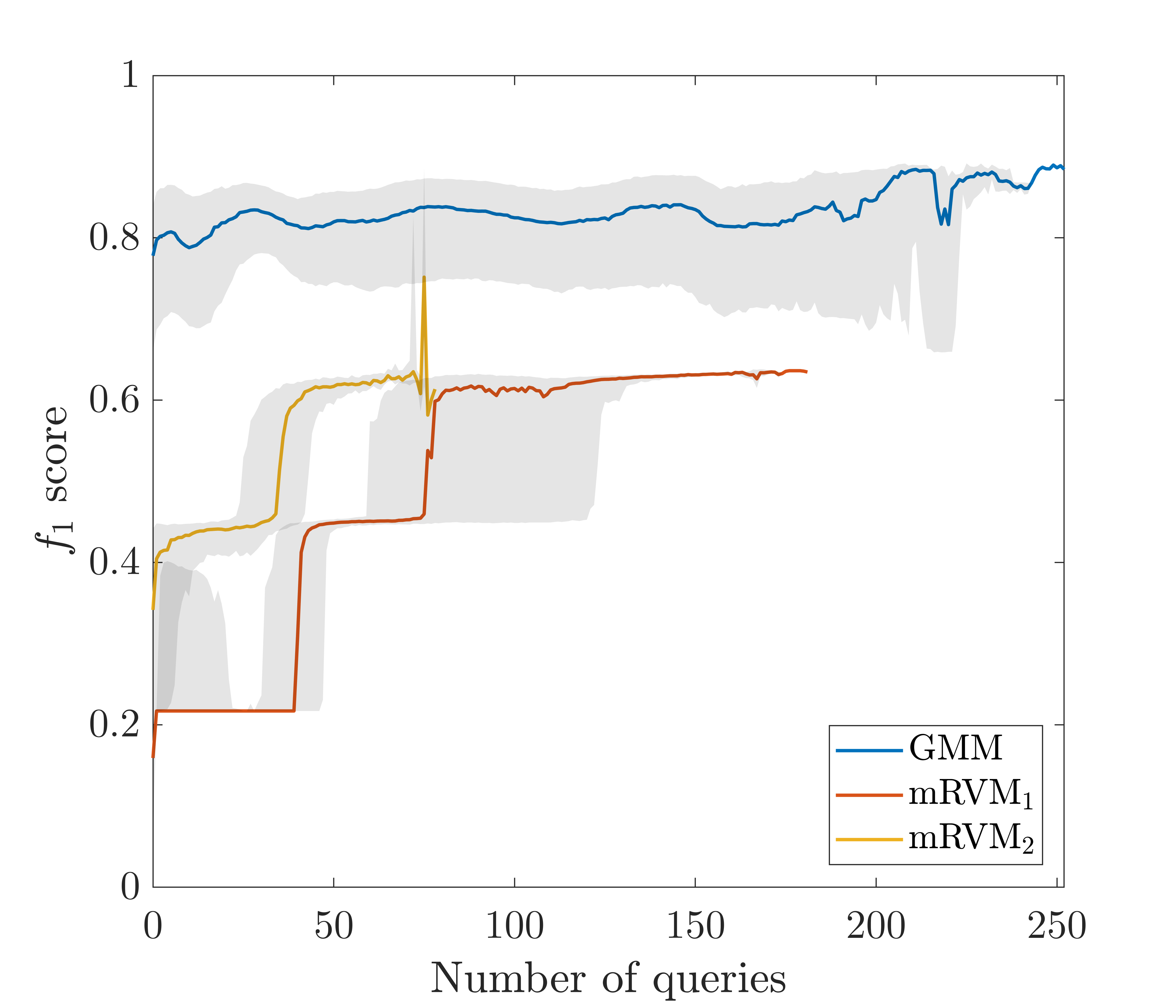}
		}
		\caption{ }
		\label{fig:f1score_rvm_z24}
	\end{subfigure}
	\caption{Variation in median (a) decision accuracy and (b) $f_{1}$ score with number of label queries for an agent utilising (i) a GMM (ii) an mRVM$_1$ and (iii) an mRVM$_2$ statistical classifier learned via risk-based active learning. Shaded area shows the interquartile range.}
	\label{fig:performance_rvm_z24}
\end{figure}

Additionally, the classification performance ($f_1$-score) is shown to highlight that the risk-based approach to active learning prioritises decision-making over classification. The decision accuracy and $f_1$-score as functions of the number of queries made are provided for the risk-based active learning of \mRVMa{} and \mRVMb{} in Figure \ref{fig:performance_rvm_z24}.

From Figure \ref{fig:dacc_rvm_z24}, it can be seen that both \mRVMa{} and \mRVMb{} surpass the GMM in terms of decision accuracy. \mRVMa{} begins with very low performance indicating that the constructive approach to selecting relevance vectors was unable to converge because of the limiting size of $\mathcal{D}_l$. Over the first few queries, however, the performance rapidly increases as $\mathcal{D}_l$ expands such that convergence of the relevance vector selection can be achieved. \mRVMb{} acheives good performances on even the limited initial $\mathcal{D}_l$. Furthermore, improvements in decision accuracy are gained at a similar, if not slightly greater, rate when compared to the GMM. This result means that superior decision-making performance is obtained throughout the querying process when adopting an \mRVMb{} classifier. It is worth noting that neither formulations of mRVM suffer from a degradation in decision-making performance over later queries - indicating robustness to sampling bias in $\mathcal{D}_l$. This result can be attributed to the discriminative nature of the models, in addition to the fact that extraneous data are excluded from $\mathcal{A}$.

\begin{figure}[ht!]
	\centering
	\scalebox{0.4}{
		\includegraphics{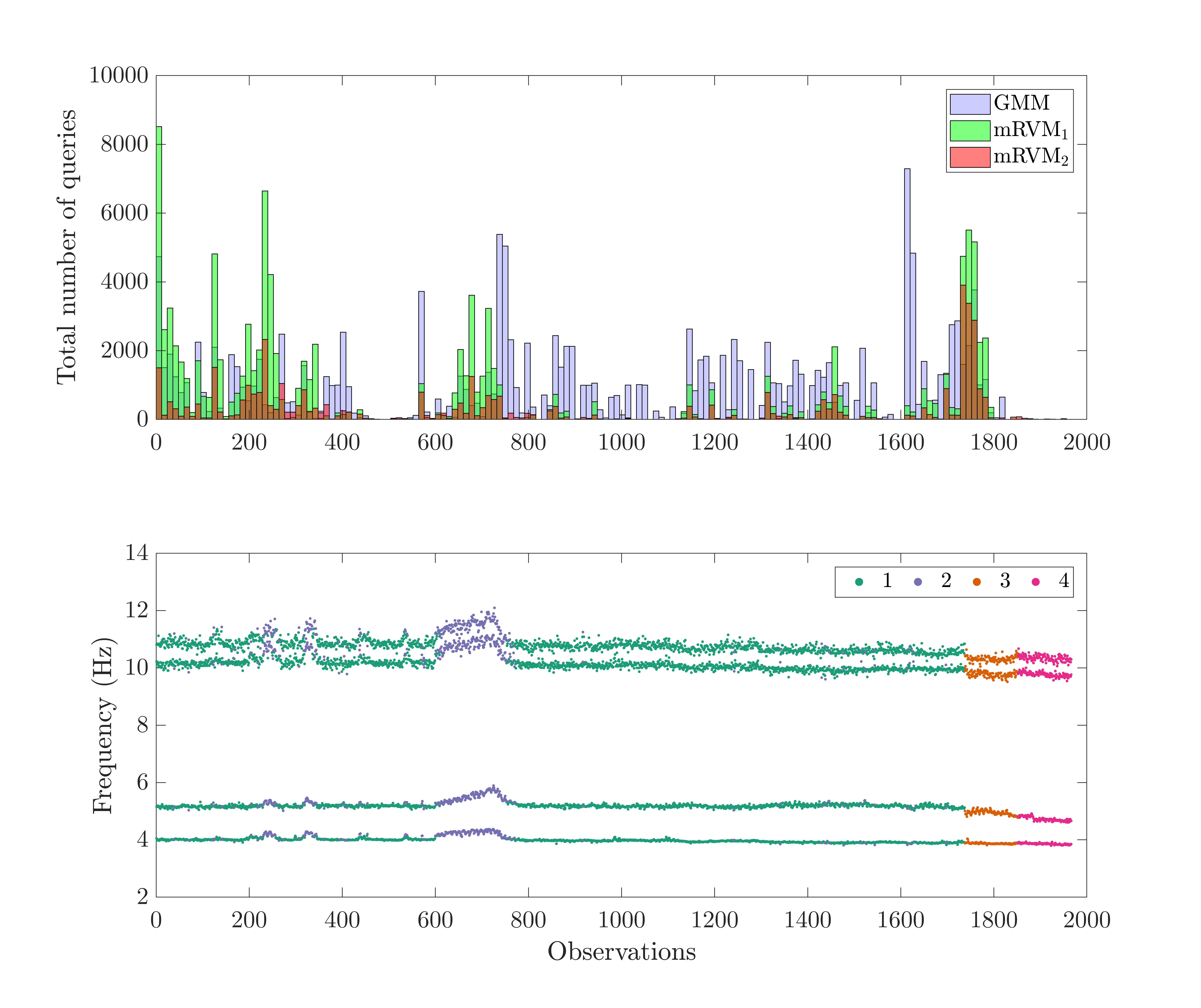}
	}
	\caption{Histograms showing the distribution of the number of queries for each observation in $\mathcal{D}_u$ from 1000 runs of risk-based active learning for (i) a GMM (blue) (ii) an mRVM$_1$ (green) and (iii) an mRVM$_2$ (red) statistical classifier. A visualisation of $\mathcal{D}_u$ is provided for reference.}
	\label{fig:all_queries_rvm_z24}
\end{figure}

From Figure \ref{fig:f1score_rvm_z24}, it can be seen that the RVMs have inferior $f_1$-scores in comparison to the GMM. Distinct improvements in classification performance arise in correlation with appearance of new classes in $\mathcal{D}_u$ (see Figure \ref{fig:all_queries_rvm_z24}). The performances plateau at approximately 0.6 as data corresponding to Class 4 are not queried, resulting in the classifiers struggling to discriminate between Classes 3 and 4. Nonetheless, as these classes share an optimal policy with respect to the decision process, decision-making performance in not impacted by these misclassifications -- as demonstrated in Figure \ref{fig:dacc_rvm_z24}.

\begin{figure}[ht!]
	\centering
	\scalebox{0.4}{
		\includegraphics{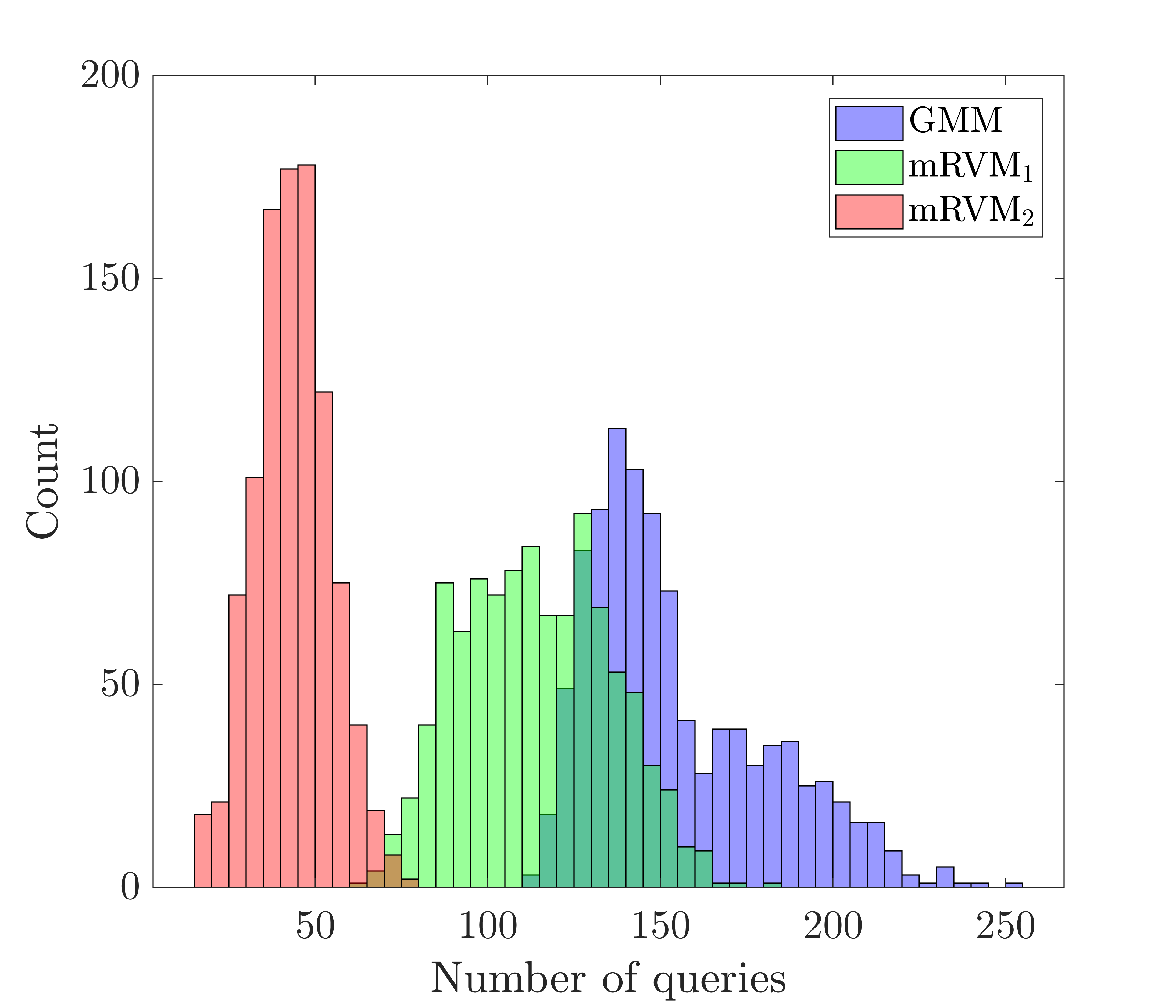}
	}
	\caption{Histograms showing the distribution of the number of queries from 1000 runs of the risk-based active learning of (i) a GMM (blue) (ii) an mRVM$_1$ (green) and (iii) an mRVM$_2$ (red) statistical classifier.}
	\label{fig:hist_rvm_z24}
\end{figure}

The distributions for the number of queries made by each active learner are shown in Figure \ref{fig:hist_rvm_z24}. \mRVMa{} and \mRVMb{} both query fewer times on average when compared to the GMM. This result can also be explained by the RVM's ability to represent classes of data using only a few prototypical samples. Between \mRVMa{} and \mRVMb{}, \mRVMb{} yields superior performance in terms of the number of queries made, with both lower mean and variance. This result is highly significant as, in the context of SHM decision processes, fewer queries mean a reduced expenditure on inspections, with lower variance indicating more consistent performance. This result demonstrates that careful selection of classifiers can save resource over the course of a monitoring campaign.

Finally, Figure \ref{fig:all_queries_rvm_z24} shows the total number of queries for each observation index in $\mathcal{D}_u$. As with Figure \ref{fig:hist_rvm_z24}, it is apparent in Figure \ref{fig:all_queries_rvm_z24} that, overall, the use of RVMs as a statistical classifier reduces the total number of queries made. It can be seen that \mRVMa{} results in increased querying early in the dataset; this is to be expected, as it was previously determined that this formulation struggles to construct a model capable of discriminating between classes, resulting in high EVPI for all regions of the feature space. As with the GMM, both RVM approaches result in an increase in the number of queries made as new patterns in the data present; specifically around observation indices 200, 700, and 1800 in $\mathcal{D}_u$. Again, this result is in agreement with the trends in the $f_1$-score shown in Figure \ref{fig:f1score_rvm_z24}. The relative reduction in the number of queries made between observations 800 to 1600, for the RVMs compared to the GMM, can be attributed to the sparse representation achieved by the RVMs. As data for these classes had already been obtained, the EVPI associated with the classes was sufficiently low such that inspection was not necessary. These results indicate that resource for inspections is expended more sparingly when using mRVMs and can be explained by the models sparse nature; as only a few data points are required to effectively represent the classes necessary for strong decision-making performance.

\section{Conclusion}

In summary, utilising probabilistic discriminative classifiers learned via risk-based active learning in decision processes  results in improved decision-making performance and improved robustness to sampling bias. The current paper used mRVMs to classify data from the Z24 bridge into various health states. These predictions were used to inform an O\&M decision process. Notably, it was demonstrated that the number of queries made throughout the learning process is decreased via the introduction of discriminative classifiers over generative models. Moreover, superior decision accuracy was achieved, indicating fewer erroneous actions were selected. These benefits are obtained by virtue of the RVM's characteristic sparsity, in addition to the reduced reliance on assumptions regarding the distribution of data. The results have significant implication in the context of decision-supporting SHM systems as they indicate that careful selection of classification models can result in lower expenditure by reducing the number of inspections required and the number of erroneous decisions made over the course of a monitoring campaign.


\section*{Acknowledgments}

The authors would like to acknowledge the support of the UK EPSRC via the Programme Grant EP/R006768/1. KW would also like to acknowledge support via the EPSRC Established Career Fellowship EP/R003625/1. LAB was supported by Wave 1 of The UKRI Strategic Priorities Fund under the EPSRC Grant EP/W006022/1, particularly the \textit{Ecosystems of Digital Twins} theme within that grant and The Alan Turing Institute. The authors would like to thank Dr.\ Robert Barthorpe of the University of Sheffield for providing valuable discussion.

	
	\bibliographystyle{splncs04}
	\bibliography{IWSHM2021_RVM}

\begin{thebibliography}{10}
\providecommand{\url}[1]{\texttt{#1}}
\providecommand{\urlprefix}{URL }
\providecommand{\doi}[1]{https://doi.org/#1}

\bibitem{Bull2019}
Bull, L., Rogers, T., Wickramarachchi, C., Cross, E., Worden, K., Dervilis, N.:
  {Probabilistic active learning: An online framework for structural health
  monitoring}. Mechanical Systems and Signal Processing  \textbf{134},  106294
  (2019)

\bibitem{Damoulas2008}
Damoulas, T., Girolami, M.: {Probabilistic multi-class multi-kernel learning:
  On protein fold recognition and remote homology detection}. Bioinformatics
  \textbf{24}(10),  1264--1270 (2008)

\bibitem{DeRoeck2003}
De~Roeck, G.: {The state-of-the-art of damage detection by vibration
  monitoring: the SIMCES experience}. Structural Control Health Monitoring
  \textbf{10}(2),  127--134 (2003)

\bibitem{Hughes2021}
Hughes, A., Barthorpe, R., Dervilis, N., Farrar, C., Worden, K.: {A
  probabilistic risk-based decision framework for structural health
  monitoring}. Mechanical Systems and Signal Processing  \textbf{150},  107339
  (2021)

\bibitem{Hughes2022}
Hughes, A., Bull, L., Gardner, P., Barthorpe, R., Dervilis, N., Worden, K.: On
  risk-based active learning for structural health monitoring. Mechanical
  Systems and Signal Processing  \textbf{167},  108569 (2022)

\bibitem{Kamariotis2022}
Kamariotis, A., Chatzi, E., Straub, D.: Value of information from
  vibration-based structural health monitoring extracted via bayesian model
  updating. Mechanical Systems and Signal Processing  \textbf{166},  108465
  (2022)

\bibitem{Maeck2003}
Maeck, J., De~Roeck, G.: {Description of Z24 benchmark}. Mechanical Systems and
  Signal Processing  \textbf{17}(1),  127--131 (2003)

\bibitem{Maeck2001}
Maeck, J., Peeters, B., De~Roeck, G.: {Damage identification on the Z24-bridge
  using vibration monitoring}. Smart Materials and Structures  \textbf{10}(3),
  512--517 (2001)

\bibitem{Peeters2001}
Peeters, B., De~Roeck, G.: {One-year monitoring of the Z24-Bridge:
  environmental effects versus damage events}. Earthquake Engineering
  Structural Dynamics  \textbf{30}(2),  149--171 (2001)

\bibitem{Psorakis2010}
Psorakis, I., Damoulas, T., Girolami, M.: {Multiclass relevance vector
  machines: Sparsity and accuracy}. IEEE Transactions on Neural Networks
  \textbf{21}(10),  1588--1598 (2010)

\bibitem{Tipping2001}
Tipping, M.: {Sparse Bayesian learning and the relevance vector machine}.
  Journal of Machine Learning Research  \textbf{1},  211--244 (2001)

\bibitem{Vega2020a}
Vega, M., Todd, M.: {A variational Bayesian neural network for structural
  health monitoring and cost-informed decision-making in miter gates}.
  Structural Health Monitoring  (2020)

\end{thebibliography}

\end{document}